\documentclass[10pt,twocolumn,letterpaper]{article}

\usepackage{cvpr}              

\usepackage{graphicx}
\usepackage{amsmath}
\usepackage{algorithm}
\usepackage{algpseudocode}
\usepackage{multirow}
\usepackage[misc]{ifsym}
\usepackage{enumitem}
\usepackage{xspace}
\usepackage{xcolor,colortbl}
\usepackage{mathrsfs}
\usepackage{amsfonts}
\usepackage{acronym}
\usepackage{bbm}
\usepackage{makecell}

\usepackage{tabularx}
\usepackage{tcolorbox} 

\usepackage{pifont}
\newcommand{\cmark}{\ding{51}}%
\newcommand{\xmark}{\ding{55}}%

\makeatletter
\renewcommand{\paragraph}{%
  \@startsection{paragraph}{4}%
  {\z@}{0ex \@plus 0ex \@minus 0ex}{-1em}%
  {\normalfont\normalsize\bfseries}%
}
\makeatother

\makeatletter
\DeclareRobustCommand\onedot{\futurelet\@let@token\@onedot}
\def\@onedot{\ifx\@let@token.\else.\null\fi\xspace}

\def\eg{\textit{e.g}\onedot} 
\def\ie{\textit{i.e}\onedot} 
 
\def\etc{\textit{etc}\onedot} \def\vs{\textit{vs}\onedot}
\def\wrt{w.r.t\onedot} 

\makeatother

\definecolor{cvprblue}{rgb}{0.21,0.49,0.74}
\usepackage[pagebackref,breaklinks,colorlinks,allcolors=cvprblue]{hyperref}

\acrodef{llm}[LLM]{large language model}
\acrodef{mllm}[MLLM]{multi-modal large language model}
\acrodef{omnillm}[OmniLLM]{Omni Large Language Model}
\acrodef{videollm}[VideoLLM]{video large language model}
\acrodef{sm}[SM]{\textit{supplementary material}}

\newcommand{\model}{M4\xspace}
\newcommand{\bench}{OmniMMI\xspace}
\newcommand{\data}{M4-IT\xspace}

\def\paperID{7934} 
\def\confName{CVPR}
\def\confYear{2025}

\title{OmniMMI: A Comprehensive Multi-modal \\ Interaction Benchmark in Streaming Video Contexts}

    \makeatletter
\def\@fnsymbol#1{\ensuremath{\ifcase#1\or \text{\Letter}\or \dagger\or
   \mathsection\or \mathparagraph\or \|\or **\or \dagger\dagger
   \or \ddagger\ddagger \else\@ctrerr\fi}}
    \makeatother

\author{Yuxuan Wang$^{1,2}$, Yueqian Wang$^{2,3}$, Bo Chen$^{1,2,4}$, Tong Wu$^{1,2}$, Dongyan Zhao$^{2,3}$, Zilong Zheng$^{1,2}\,$\thanks{Corresponding author: Zilong Zheng $\langle$\texttt{zlzheng@bigai.ai}$\rangle$}\\
\small $^1$ Beijing Institute for General Artificial Intelligence\quad{} $^2$ State Key Laboratory of General Artificial Intelligence \\
\small $^3$ Wangxuan Institute of Computer Technology, Peking University\quad{} $^4$ X-LANCE Lab, Shanghai Jiao Tong University \\
{\tt\small \{wangyuxuan1, zlzheng\}@bigai.ai} \\
\small \url{https://omnimmi.github.io}
}

\begin{document}
\maketitle

\begin{abstract}
The rapid advancement of multi-modal language models (MLLMs) like GPT-4o has propelled the development of Omni language models, designed to process and proactively respond to continuous streams of multi-modal data. Despite their potential, evaluating their real-world interactive capabilities in streaming video contexts remains a formidable challenge. In this work, we introduce \textbf{OmniMMI}, a comprehensive multi-modal interaction benchmark tailored for OmniLLMs in streaming video contexts. OmniMMI encompasses over 1,121 videos and 2,290 questions, addressing two critical yet underexplored challenges in existing video benchmarks: streaming video understanding and proactive reasoning, across six distinct subtasks.  Moreover, we propose a novel framework, Multi-modal Multiplexing Modeling (M4), designed to enable an inference-efficient streaming model that can \textbf{see, listen while generating}.
Extensive experimental results reveal that the existing MLLMs fall short in interactive streaming understanding, particularly struggling with proactive tasks and multi-turn queries. Our proposed \model, though lightweight, demonstrates a significant improvement in handling proactive tasks and real-time interactions.
\end{abstract}
\vspace{-5mm}

\section{Introduction}
\label{sec:intro}

The burgeoning field of \acp{mllm}, exemplified by GPT-4o~\cite{gpt4o} and Gemini Pro~\cite{gemini}, marks a significant leap towards embodied agentic intelligence by incorporating multi-modal encoders within pre-trained \acp{llm}, such as video understanding~\cite{videollava,qwen2vl,vinvl}, audio comprehension~\cite{qwen2audio}, and speech-to-speech dialogue~\cite{xie2024miniomni,xie2024miniomni2,fang2024llama,defossez2024moshi}, \etc. The overarching goal is \textbf{(i)} to transcend the general capabilities of \acp{llm} to process and respond to \textbf{continuous streams} of multi-modal dynamics, encompassing text, vision, and speech modalities, \ie, \acp{omnillm}, 
and \textbf{(ii)} to derive \textbf{interactive systems} that can take the first-person perspective to interact with the real world. However, this rapid development raises a crucial question: \textit{How can we effectively evaluate the real-world interactive capabilities of OmniLLMs in streaming video contexts?} Addressing this question is pivotal to validating their design efficacy and enhancing their performance for comprehensive open-world multi-modal understanding.

\begin{figure*}[t!]
  \centering

      \includegraphics[width=.98\linewidth]{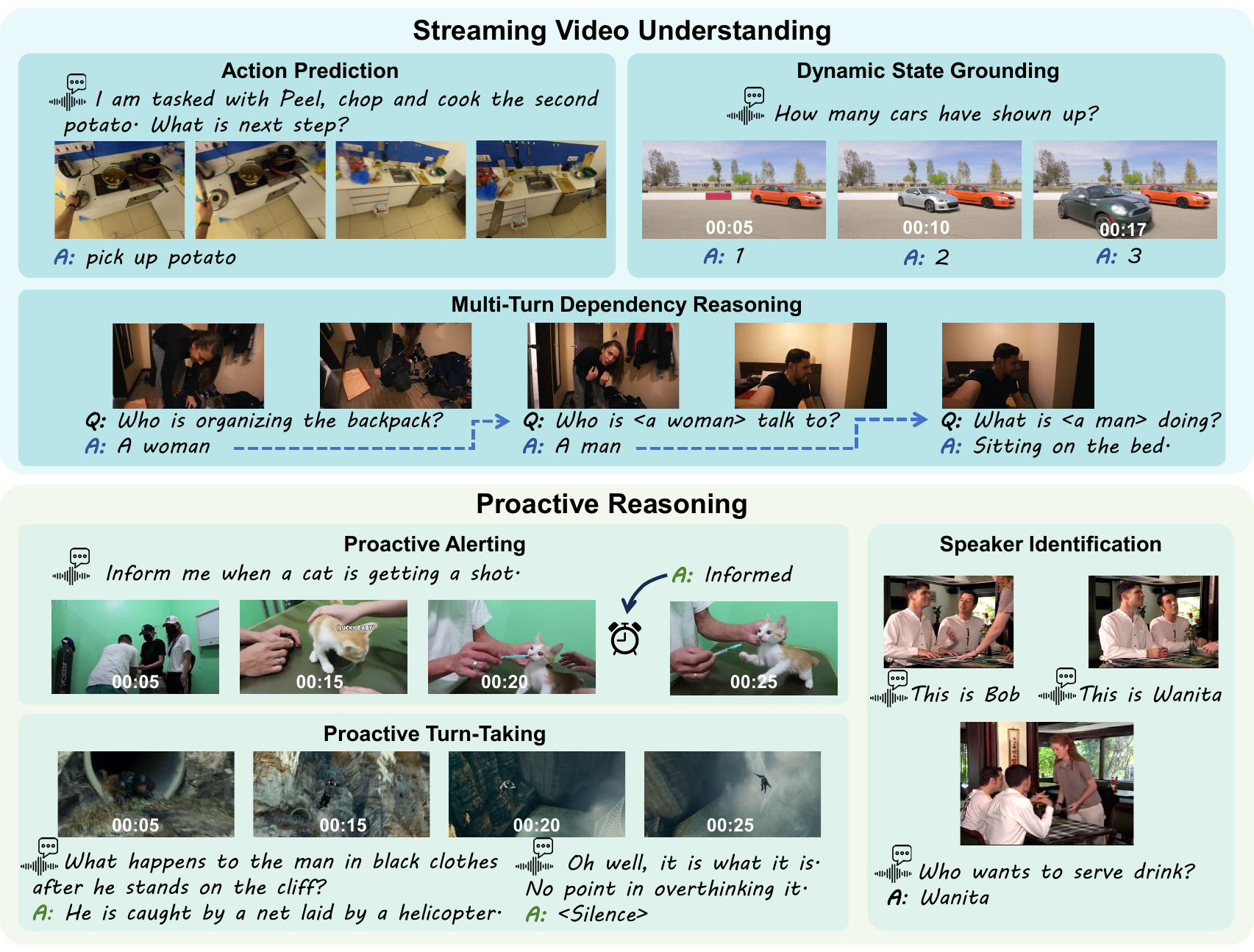}
      \vspace{-3mm}
   \caption{\bench consists of two categories of multi-modal interactive challenges: streaming video understanding (top) and proactive reasoning (bottom). Each query is processed into natural language text and synthetic audio as input. }
   \label{fig:onecol}
   \vspace{-.2in}
\end{figure*}

In response, a vast number of benchmarks has been launched with different focuses on long-form video understanding~\cite{mlvu,videomme,videoxl}, comprehensive video analysis~\cite{videomme}, or audio-video understanding~\cite{li2024omnibench}, \etc. However, most of these benchmarks take in the entire video sequence as input, some of which use frame selection techniques with slight information scarification, to produce final answers. These tasks, though challenging, are far from the real-world interactive scenarios where videos are taken in as a streaming sequence; refer to \cref{tab:comparison} for benchmark comparisons. More recently, OmniBench~\cite{li2024omnibench} has been introduced to evaluate models' capabilities on visual, acoustic, and textual inputs simultaneously. However, only the last image frame is considered visual input, while the video dynamics across streaming contexts and interactive features are overlooked.

To bridge the critical gap, we introduce \textbf{M}ulti-\textbf{m}odal \textbf{I}nteraction Benchmark for \acp{omnillm} (\textbf{\bench}), which aims at comprehensively evaluating the interactive capabilities of streaming video context in the open world (\cref{fig:onecol}). We start by formalizing the task of streaming multi-modal understanding ($\S$~\ref{sec:tasks}). Apart from challenges identified by prior long-form video-audio benchmarks (\eg, temporal dynamics~\cite{videomme,mlvu} and multi-modal localization~\cite{mmneedle,videollamb}), \bench considers two featured obstacles for real-time interactions. 
\begin{itemize}[leftmargin=*, noitemsep, topsep=0pt]
    \item \textbf{Streaming Temporal State Awareness.}  Streaming video understanding must build an understanding \wrt the current and historical temporal state incrementally, without accessing the future context. This contrasts with traditional \acp{mllm} that can leverage the entire multi-modal contexts, posing challenges in our distinguished tasks of action prediction (AP), state grounding (SG) and multi-turn dependencies (MD) ($\S$\ref{sec:streaming}).
    \item \textbf{Proactive Reasoning and Turn-Taking.} Generating responses proactively and appropriately anticipating the turn-taking time spot \wrt user's intentions and dynamic contexts is a crucial feature for general interactive agents. This typically requires models to identify speakers (SI), distinguish between noise or legitimate query (PT), and proactively initiate a response (PA) ($\S$\ref{sec:proreason}).
\end{itemize}

In light of this, \bench is crafted as the first-ever comprehensive \ac{omnillm} benchmark to address the aforementioned streaming and interactive challenges. As exemplified in \cref{fig:onecol}, we curate a dataset of 1,121 videos sourced from YouTube and open-sourced video-audio data with an average duration of 324 seconds. 2,290 questions are manually annotated and reviewed \wrt both visual and auditory information from video inputs, encompassing different topics relevant to the multi-modal contexts~(\cref{fig:category}). Notably, to enhance the interactive feature of this benchmark, we designed multi-turn questions (up to 3 turns) where the next response is based on the answer from previous turns; see dynamic state grounding and multi-turn dependency reasoning as examples.  

Using \bench, we make a thorough evaluation of various well-known \acp{videollm} and accessible \acp{omnillm} across all tasks. Surprisingly, most models encounter challenges in multi-turn tasks involving streaming video, where they struggle beyond a single reasoning step, thus revealing limitations in dynamic settings. In the realm of audio-visual interaction, models that process both audio and visual inputs do not outperform those that handle visual inputs alone, indicating inadequate modality alignment. Furthermore, increasing the model size does not lead to improved performance; instead, models capable of processing longer input lengths demonstrate superior results, highlighting the importance of balancing input length with memory efficiency.


\begin{table*}[t!]
    \vspace{-.1in}
    \small
    \caption{\textbf{Comparison with existing Video Benchmark} \bench is the first comprehensive benchmark that focuses on streaming and interactive VideoLLMs. * only the last frame of videos are taken as input. \textsuperscript{\dag} is not a real-time setting.}
    \label{tab:comparison}
    \centering
    \resizebox{\linewidth}{!}{
    \begin{tabular}{lccc *{2}{>{\centering\arraybackslash}p{1cm}} ccccc}
        \toprule

        \multirow{2}{*}{\textbf{Benchmark}} & \multirow{2}{*}{\textbf{Modality}} & 
        \multirow{2}{*}{\textbf{\#Videos}} &\multirow{2}{*}{\textbf{\#Questions}} &\multicolumn{2}{c}{\textbf{Length}} & 
        \multirow{2}{*}{\makecell{\textbf{Multi-Hop}}} & \multirow{2}{*}
        {\makecell{\textbf{Contain-Ego}}}  & 
        \multirow{2}{*}{\textbf{Streaming}} & \multirow{2}{*}{\makecell{ \textbf{Proactive}}} & \multirow{2}{*}{\textbf{Interactive}}   \\ 
        \cmidrule{5-6} & & & & \textbf{Video(s)} & \textbf{\#Turn}  \\\midrule
        MVBench~\citep{mvbench} & Video  & 3,641 & 4,000 & 16.0 & 1  & \xmark & \cmark & \xmark & \xmark & \xmark  \\
        Video-Bench~\citep{videobench} & Video  & 5,917 & 17,036 & 56.0 & 1  & \xmark & \xmark & \xmark & \xmark & \xmark \\
        EgoSchema~\citep{egoschema} & Video  & 5,063 & 5,063 & 180.0 & 1  & \xmark & \cmark & \xmark & \xmark & \xmark \\
        VideoMME~\citep{videomme} & Video, Audio  & 900 & 2,700 & 1017.9  & 1& \xmark & \xmark & \xmark & \xmark & \xmark \\
        LongVideoBench~\citep{longvideobench} & Video  & 3,763 & 6,678 & 473.0  & 1 & \xmark & \xmark & \xmark & \xmark & \xmark \\
        MLVU~\citep{mlvu} & Video  & 2,593 & 2,593 & 720.0 & 1  & \xmark & \cmark & \xmark & \xmark & \xmark \\
        MultiHop-EgoQA~\citep{multihop-ego} & Video  & 360 & 1,080 & 180.0 & 1  & \cmark & \cmark & \xmark & \xmark & \xmark \\
        OmniBench~\citep{li2024omnibench} & Image*,Audio  & - & 1,142 & 9.2 & 1  & \xmark & \cmark &\xmark & \xmark & \xmark \\   
        StreamingBench~\citep{streamingbench} & Video, Audio  & 900 & 4,500 & 243.1 & 1  & \xmark & \xmark &\cmark & \cmark\textsuperscript{\dag} & \xmark \\         
        OvO-Bench~\citep{ovobench} & Video  & 644 & 2,814 & 428.9 & 1  & \xmark & \cmark &\cmark & \cmark\textsuperscript{\dag} & \xmark \\         \midrule

        \bench (Ours) & Video, Audio  & 1,121   & 2,290  & 324.3   &1-3& \cmark & \cmark & \cmark & \cmark & \cmark \\
         \bottomrule
    \end{tabular}
    }    
    \vspace{-.2in}
\end{table*}

To margin towards \textit{real-time} interactive reasoning, in $\S$\ref{sec:model}, we devise a novel and robust framework, \textbf{M}ulti-\textbf{m}odal \textbf{M}ultiplexing \textbf{M}odeling (\textbf{\model}), taking inspirations from duplexing modeling of auditory models~\cite{lslm,murahari2022datamux,xie2024miniomni}. We crafted a small video-free SFT data, \data, for proactive turn-taking awareness. As such, \model can be built upon any pre-trained VideoLLMs, enabling an inference-efficient streaming model that can \textbf{see, listen while generating}.






\section{Related Work}

\paragraph{Omni Large Language Models}
The advancement of \acp{omnillm} represents a notable achievement in \acp{mllm}, aiming for real-time comprehension and processing of diverse modalities. Significant contributions in this domain include GPT-4o and Project Astra, which manage and generate multi-modal inputs and outputs encompassing text, audio, images, and videos. Concurrently, the open-source community has developed models like VITA~\cite{vita} and Ocean-Omni~\cite{li2024baichuanomni}, which integrate distinct models for enhanced non-awakening and interrupting capabilities. Additionally, audio-based conversational models such as  LSLM~\cite{lslm} and mini-Omni~\cite{xie2024miniomni,xie2024miniomni2} have emerged, utilizing text-instructed speech generation to facilitate real-time speech interactions while maintaining strong language proficiency. Despite these advancements, existing models often lack proactive reasoning abilities (\cref{sec:proreason}) without additional computational overhead. Addressing this gap, we introduce \model, an interactive framework for \acp{mllm} that enables proactive reasoning without necessitating extra forward computation steps or video-specific training.

\paragraph{Video Understanding Benchmarks}

Early general video understanding benchmarks~\cite{msrvtt_qa, tgif, anet, mvbench, videobench, videohallucer} were introduced to evaluate models' capabilities in general video-language understanding. Other works~\cite{star,nextqa} have focused on temporal grounding for dynamic video content. More recently, several benchmarks for long video understanding ~\cite{egoschema, videomme, longvideobench, mlvu} have been proposed. The MultiHop-EgoQA dataset ~\cite{multihop-ego} introduces a multi-hop video question-answering dataset with temporal evidence to assess models' multi-hop reasoning abilities over relatively long videos. OmniBench ~\cite{li2024omnibench} extends visual information to include audio, proposing an audio-focused benchmark complemented by visual information. However, there is a lack of comprehensive benchmarks for streaming video understanding and proactive reasoning for interactive \acp{videollm}. To address this gap, we propose \bench, aiming to encourage further advancements in this area.

\paragraph{Streaming Video Understanding}

The realm of \acp{mllm}~\cite{qwen2vl,internvideo2,videoxl,longva}
have achieved superior performance in various video-centric tasks by employing a progressive training paradigm that unifies different self- or weakly-supervised learning frameworks. 
Some recent \acp{videollm}~\cite{mallm,streamingvideo,streamingcap,streaminglv,videollamb} enables video processing in an online manner and store past video information in a memory bank, facilitating long-term video analysis without exceeding computational constraints. VideoLLM-online~\cite{videollmonline} further addresses the challenge of integrating diverse data modalities by effectively introducing a special token after each frame with a binary classification task. These advancements underscore the transformative potential of multimodal and interactive video understanding technologies, promising innovative applications across various fields as models continue to improve in their ability to integrate and process diverse data modalities.

\section{The \bench Dataset}
\label{sec:bench}

   \begin{figure}[t!]
   \includegraphics[width=.9\linewidth]{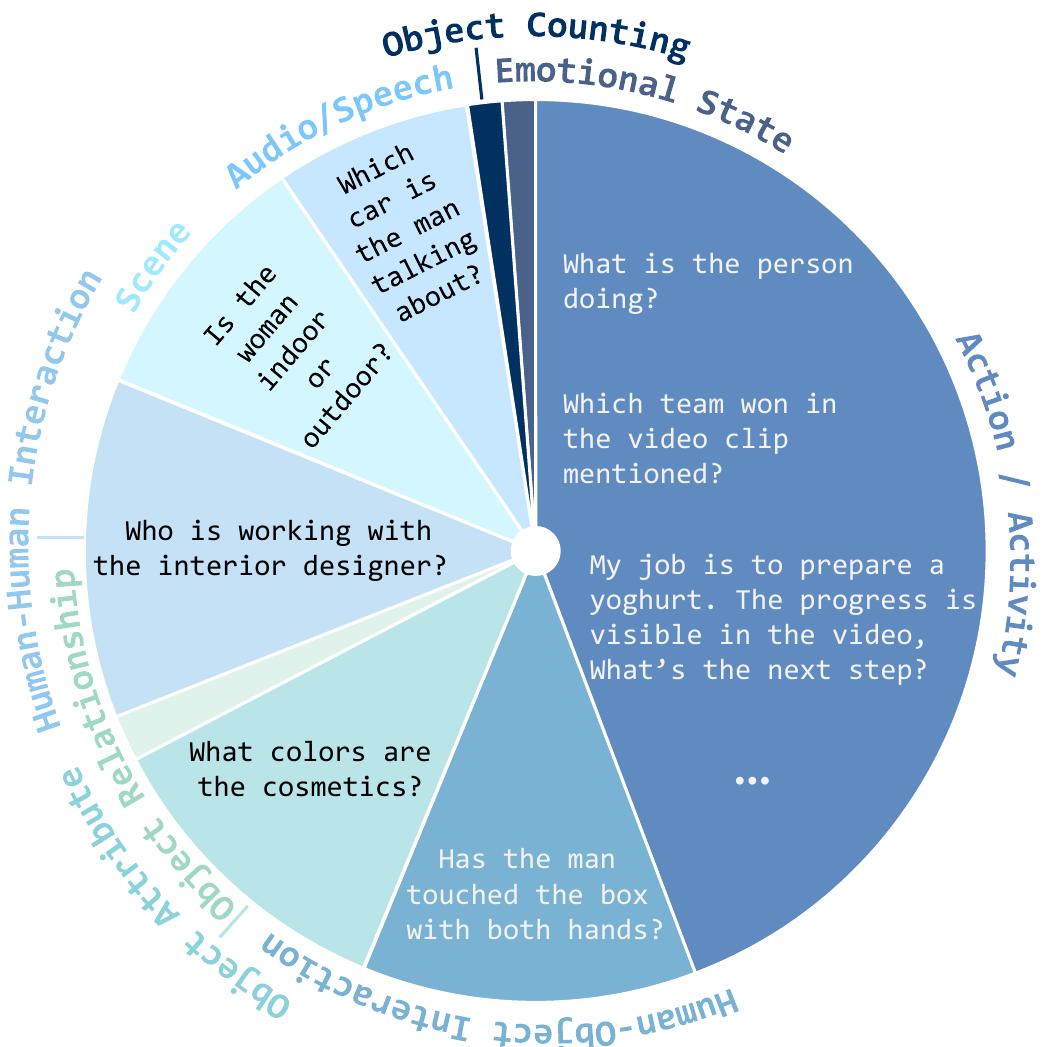}
   \captionof{figure}{\small Distribution and examples of different types of query prompts.}\label{fig:category}
   \vspace{-.2in}
   \end{figure}

\subsection{Dataset Construction}

\paragraph{Data Source} 
The existing datasets include Ego4d~\cite{ego4d}, COIN~\cite{coin}, Shot2Story20K~\cite{shot2story20k}, QVHighlight~\cite{qvhighlight}, and MLVU~\cite{mlvu}, which encompass both egocentric and open-domain videos across diverse topics. However, most of these datasets are not specifically aligned with typical interactive scenarios, such as interactions involving a camera. To address the issue and minimize data contamination, we integrated the test and validation sets from the aforementioned datasets with newly collected video footage. Specifically, we augmented these datasets by sourcing additional videos from YouTube, utilizing 425 keywords pertinent to frequently observed interactive environments. For each keyword, we downloaded a maximum of 50 videos, each with a duration of less than one hour. Ultimately, we conducted a manual review to filter out low-quality videos, resulting in a curated set of 78 videos. A comprehensive list of these keywords is provided in the \ac{sm}.

\paragraph{Annotation}

The annotation process requires annotators to analyze both visual and auditory elements of input videos. Initially, annotators review the video chronologically and annotate the time span of relevant actions or states, tailored to the specific task, particularly for streaming video understanding. We provide question-type prompts to guide annotators in focusing on various video aspects, such as actions or object attributes, detailed in \ac{sm}. For tasks involving proactive alerting and turn-taking, annotators record timestamps for significant events or their conclusions. In speaker identification tasks, annotators mark time spans of human introductions and subsequently label individuals' names following activities or special situations. To ensure benchmark quality, annotators must review the video again, considering the noted spans and original dataset information, to refine questions and answers based on their annotations.

\paragraph{Quality Review}
To ensure accuracy, a second annotator reviews the initial annotations, focusing on the consistency of the questions, answers, and time spans. Any inconsistencies identified are documented and corrected. We then calculate the inter-annotator agreement to evaluate the consistency and reliability of the annotations. We re-used VIA tool\footnote{\scriptsize \url{https://www.robots.ox.ac.uk/~vgg/software/via}} for the annotation and reviewing process. A detailed description of the annotation procedure and the annotation interface is provided in \ac{sm}.

\subsubsection{Statistics}


In \cref{tab:comparison}, we present a comparison of our benchmark with existing popular benchmarks in the field of video understanding. Our dataset consists of 1,121 videos and 2,290 queries, with the average video length exceeding five minutes. Notably, a segment of our benchmark includes multiple turns, requiring models to correctly answer all associated questions to be labeled as a hit. This approach simulates a streaming scenario, thereby introducing an additional challenge. Overall, our work is the first to provide a comprehensive benchmark specifically designed to evaluate the efficacy of streaming video understanding models. \cref{tab:statistics} presents comprehensive statistics of the videos and queries across all six subtasks. We further delineate topics of query prompts into different categories to demonstrate the diversity of questions. The distribution and examples are shown in \cref{fig:category}. As seen, queries related to action/activity predominates in our benchmark, reflecting the dynamic nature of streaming video.

   \begin{table}[t!]
       \centering
    \small
      \captionof{table}{  \bench detailed statistics. Vid.(s): video duration. Que.: question words. }\label{tab:statistics}
    \setlength{\tabcolsep}{3pt}

   \resizebox{\linewidth}{!}{
    \begin{tabular}{lc*{3}{>{\centering\arraybackslash}p{1cm}}*{3}{>{\centering\arraybackslash}p{1cm}}}
        \toprule

        \multirow{2}{*}{\textbf{Statistic}} & \multicolumn{3}{c}{\textbf{Streaming}} &\multicolumn{3}{c}{\textbf{Proactive}}  \\ \cmidrule{2-4}
        \cmidrule{5-7} &\textbf{SG} & \textbf{AP} & \textbf{MP} &\textbf{PT} & \textbf{PA} & \textbf{SI}  \\\midrule
        \#Videos & 300 & 200 & 300  & 78 & 200  & 200  \\
        \#Queries & 704 & 200 & 786  & 200 & 200  & 200  \\
        Avg. Turns & 2.35 & 1.00 & 2.62 & 1.00 & 1.00  & 1.00  \\
        Avg. Vid.(s) & 350.82 & 234.95 & 374.80  & 2004.10 & 149.82  & 549.64  \\
        Avg. Que. & 16.00 & 25.99 & 26.27 & 8.45 & 17.49  & 60.91  \\
         \bottomrule
    \end{tabular}
    }  
   \end{table}

   \begin{figure}[t!]
       \centering
       \includegraphics[width=\linewidth]{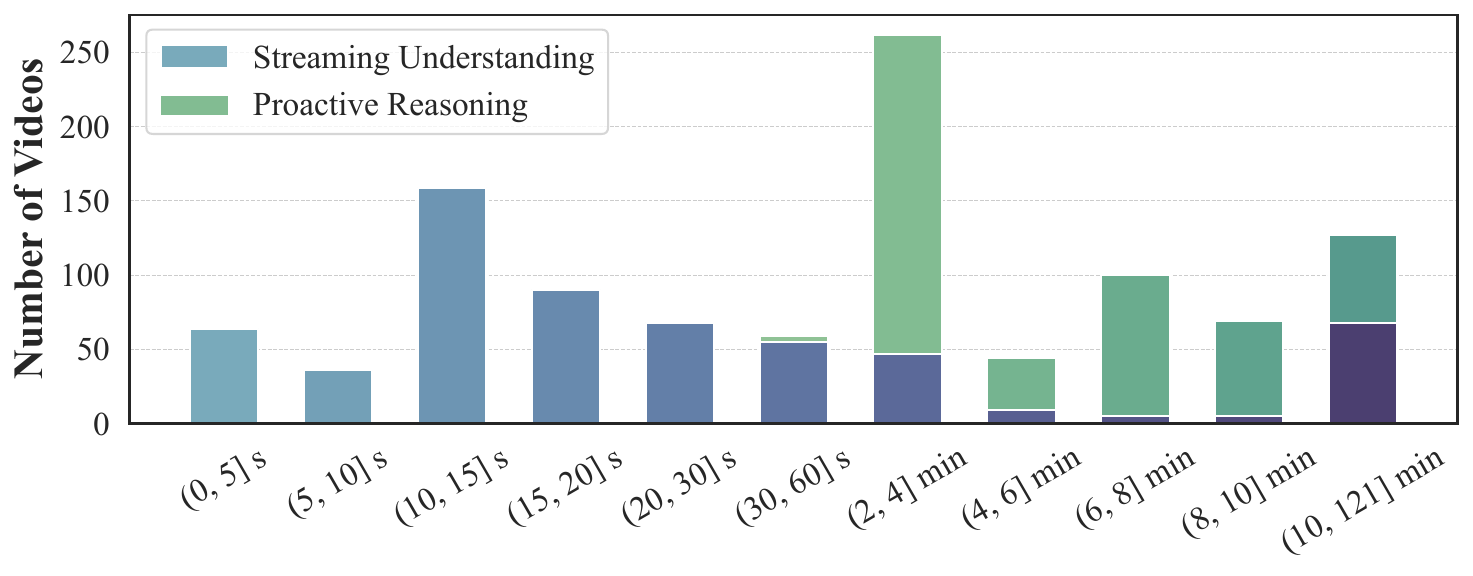}
              \vspace{-8mm}
       \caption{Distribution of video duration length.}
       \label{fig:enter-label}
           \vspace{-.16in}
   \end{figure}

\subsection{Benchmark Tasks}\label{sec:tasks}

\paragraph{Streaming Video Formulation}We formalize an input streaming video as an infinite video sequence $V_\infty = \{v_1, \cdots, v_t, \cdots\}$, where $v_t$ corresponds to the temporal time $t$ and $V_t$ is the video history up to time $t$. $\Delta t$ is the minimum temporal unit that denotes the interval between consecutive frames. At timestamp $t$, let $q_t^n$ denote the $n$-th natural language user query in the form of text or audio, $H^n_t=\{(q^1_t, a^1_t), \cdots, (q^{n-1}_t, a^{n-1}_t)\}$ denote the interaction history prior to $t$, the task is to generate next response \wrt the input streaming context, \ie, $a_t^n = f(V_t, H^n_t, q^n_t)$, where $a_t^n$ is the $n$-$th$ predicted answer at time $t$, Of note, $a_t$ could be an action to be executed in a short-term future; see \cref{sec:proreason} for exemplar tasks.


\subsubsection{Streaming Video Understanding}\label{sec:streaming} 


\paragraph{Dynamic State Grounding (SG):} This task aims to ascertain the dynamic states of a streaming video at different timestamps. We repeatedly pose the same query at different temporal states, \ie, $\{q_\tau = q_t; \tau \sim [1, t)\}$. The objective is to determine the correct answer $a^i_\tau$ for each timestamp, where $a^i_\tau$ depends on $V_{\tau-\mu \Delta t}$, with $\mu \Delta t$ being a short duration preceding the time $\tau$.


\paragraph{Action Planning (AP):} Given a natural language goal and a historical sequence from the streaming video, this task is to identify the correct next action to achieve the goal.


\paragraph{Multi-turn Dependency (MD) Reasoning:} This task involves answering a series of questions where each subsequent question depends on the answer to the previous one. The requirement is that $a^i$ forms a part of $q^{i+1}$, \ie, $a^i = F(q^i, \{a^1, a^2, \ldots, a^{i-1}\})$ and $q^{i+1} = G(q^i, a^i)$, where $G(\cdot,\cdot)$ generates the next question based on the current answer in a predefined question template. The correctness of each answer $a^i$ is contingent upon the accuracy of the preceding answers.

\noindent\textit{\underline{Metrics}}\quad We use GPT4-o to assess each step of \textbf{SG} and \textbf{MD}, and the final accuracy is averaged over the correct responses to all questions in the sequence. For \textbf{AP}, the evaluation is computed as the accuracy of selecting a response from a predefined vocabulary set of action candidates.

\begin{figure*}[t!]
  \centering
   \includegraphics[width=.92\linewidth]{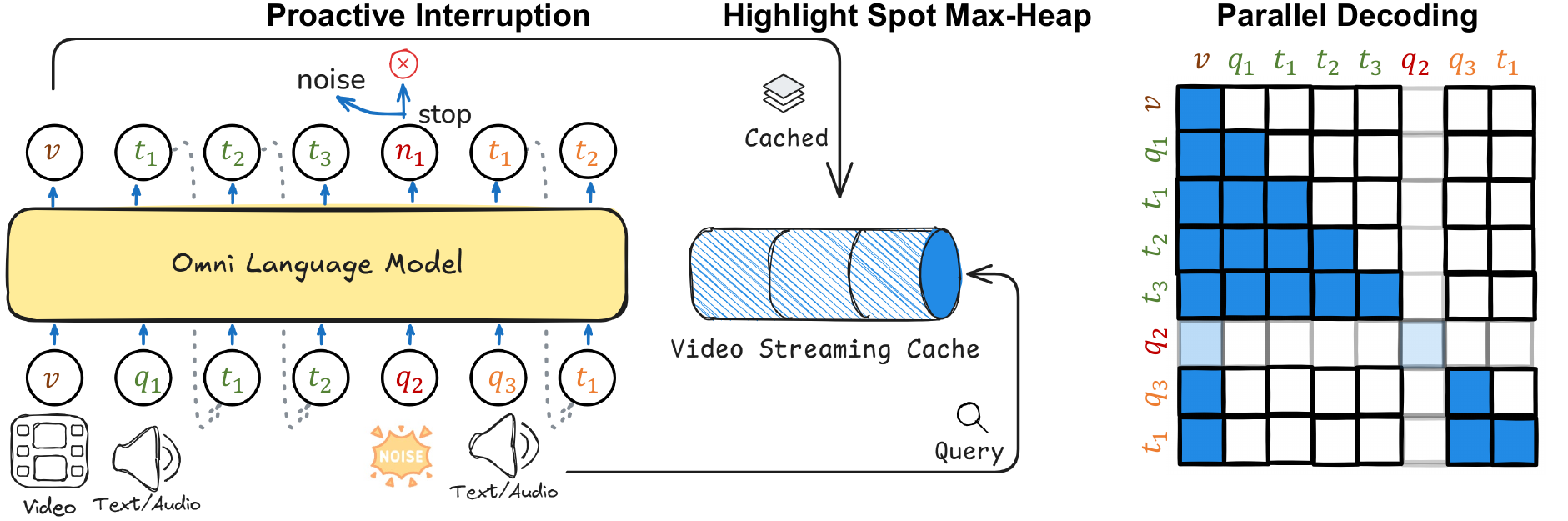}
   \caption{Multiplexing Modeling of \model. $v$ is the streaming video, $q_i$ denotes the input query, $t_i$ indicates the generated token, $n_i$ denotes noise token which will be discarded from the KVCache. The streaming video KVCache is computed to trigger a highlight spot index for the next response generation. Proactive interruption is facilitated through the computation of specific tokens designed for noise and stop signals.  The parallel decoding takes mask strategy with dynamic KVCache to process multiple queries in one forward step. }
   \label{fig:parallel}
   \vspace{-.2in}
\end{figure*}

\subsubsection{Proactive Reasoning}\label{sec:proreason}


\paragraph{Speaker Identification (SI):} In streaming videos featuring multiple individuals, a proficient model should accurately identify speakers to better comprehend multi-party dialogues. Given an introduction by oneself or others, the question $q^i_t$ pertains to the current situation and requires identifying the name of the corresponding speaker.

\paragraph{Proactive Alerting (PA):} A critical application of streaming video understanding is in surveillance, where the model is expected to notify humans of potentially dangerous situations. The desired response is a proactive alerting function $a^t=A(v)$ to be executed on the consecutive video sequences until appropriate altering information (\eg, ``\texttt{informed}'' in \cref{fig:onecol}) is proactively announced.

\paragraph{Proactive Turn-taking (PT):} Streaming videos often contain significant noise. A competent model should distinguish between queries that require a response and those that are merely noise, necessitating silence. We construct a series of queries that do not require a response to evaluate the model's ability to resist responding to noise queries. 

\noindent\textit{\underline{Metrics}}\quad We employ GPT4-o to evaluate the accuracy of the \textbf{SI} metric. For \textbf{PA}, the timestamp of the model's initial proactive response is recorded and considered a successful instance if it occurs within the designated timeframe. For \textbf{PT}, accuracy is determined by calculating the percentage of instances where no response is generated.

\section{Multi-modal Multiplexing Modeling}\label{sec:model}

We evaluate a number of popular open-source  \acp{mllm} on \bench in \cref{tab:overall}. Surprisingly, the
existing MLLMs are far from satisfactory in streaming video understanding. To fill the gap, we develop a robust
OmniLLM baseline, \textbf{M}ulti-\textbf{m}odal \textbf{M}ultiplexing \textbf{M}odeling, which is dubbed as \textbf{\model}.

Motivated by recent advances in speech LMs~\cite{lslm,xie2024miniomni}, \model formulates the interactive challenges of real-time multi-modal communications with multiplexing modeling~\cite{murahari2022datamux}, a technique that enables LMs process multiple inputs simultaneously with a single compact representation. Compared with traditional VideoLLMs and OmniLLMs, \model presents advances in:
\begin{itemize}[leftmargin=*, topsep=0pt, noitemsep]
    \item \textit{Proactive Generation.} A critical aspect of streaming video understanding is the model's ability to proactively generate the next response without human intervention. 
    \item \textit{Proactive Interruption.} When presented with a new query, \model determines whether it is \textbf{legitimate} or merely \textbf{noise} in a single forward step. 
    \item \textit{Efficient Parallel Decoding.} With multiplexing inputs, \model decodes the next token in parallel to the inputs.
\end{itemize}



\paragraph{Proactive Generation}
Most current methodologies~\cite{videollmonline, vita, streamingcap} employ special tokens in conjunction with binary classification tasks and threshold settings to enable continuous narration. However, the efficacy of these special tokens is heavily contingent upon the chosen threshold settings, leading to significant obstacles for generalizing across various domains and video-language models. Building on insights from recent works~\cite{gpt4o,xie2024miniomni2}, we ask: \textit{whether it is possible to enable real-time proactive generation without supplementing time-consuming video-specific training?}

\noindent\textit{\underline{Solution}}\quad We answer the question with affirmation. In \model, we derive an attention-based inference method, namely highlight spot, by harnessing the potential within pre-trained VideoLLMs. Given a streaming video $V_\infty$ and a query $q_t$ as in \cref{sec:tasks}, the algorithm goes as follows; refer to \ac{sm} for a formalized pseudo-code:

\noindent$\rhd$ \textbf{Step I: Streaming KV Cache.} For each in-coming frame $v$, we pre-compute $K=W_kv$ and $V=W_V v$ vectors to form a KV cache.
The attention scores between the query $q$ and the frames are calculated \wrt the KV cache, \ie, $s = \text{softmax}\left(\frac{qK^{\rm T}}{\sqrt{d_k}}\right)$, with their mean and variance as $\mu,\sigma$.

\noindent$\rhd$ \textbf{Step II: Highlight Spot Max-heap.} Indices of frames whose attention scores exceed the Gaussian average $\mu + \alpha \times \sigma$ are stored in a max heap, where $\alpha$ is a Gaussian factor.

\noindent$\rhd$ \textbf{Step III: Hit Computation.} The peak index from max-heap is extracted. If a frame index has a higher occurrence frequency than a predetermined threshold, it is designated as an ``\texttt{alert}'', triggering a response generation.



\paragraph{Interruption Detection}
\textit{\underline{Starting Detection}} In this process, we calculate the probability of the ``\texttt{<bos>}'' token as a reference point. Drawing inspiration from \citet{medusa}, we utilize the reciprocal of perplexity as the threshold for identifying this special token.
\begin{align}
\small
    &p(x_{n+k} \mid x_1, x_2, \ldots, x_{n+k-1}) > \nonumber \\ &\beta \cdot \exp \left( -S\left( p(\cdot \mid x_1, x_2, \ldots, x_{n+k-1}) \right) \right),
\end{align}
where $\beta$ is a scaling factor, $S(\cdot)$ is the entropy function. The threshold for noise detection is dependent on the perplexity of the model. When there is a larger perplexity, the threshold is reduced, indicating the query is more like a noise that does not need a response.

\noindent\textit{\underline{Stopping Detection}} Knowing when to stop is a critical feature of an interactive system, which we consider essential for developing a duplex system. Similar to noise detection, when presented with a new query, we assess whether to halt the generation process by calculating the probability of the ``\texttt{<eos>}'' token in a single forward pass. This decision is made using the same threshold employed in noise detection.

\begin{table*}[t!]
        
    \caption{\textbf{Performance comparison of existing \acp{videollm} on \bench}.  The 1st, 2nd, 3rd of \textbf{SG} and \textbf{MD} tasks represent the cumulative accuracy up to and including these stages. The ``avg.'' indicates average accuracy across all data points. } 
    \label{tab:overall}
    \centering
    \resizebox{\linewidth}{!}{
    \begin{tabular}{l|c|c|ccccccccc|ccc}
        \toprule
        \multirow{2}{*}{\textbf{Models}} & \multirow{2}{*}{\textbf{LLM}} & \multirow{2}{*}{\makecell{\textbf{Num} \\ \textbf{Frames}}} & \multicolumn{4}{c}{\textbf{SG}} & \multirow{2}{*}{\textbf{AP}} & \multicolumn{4}{c|}{\textbf{MD}} & \multirow{2}{*}{\textbf{SI}} & \multirow{2}{*}{\textbf{PA}} & \multirow{2}{*}{\textbf{PT}}  \\ 
        \cmidrule(lr){4-7} \cmidrule(lr){9-12}
        & & & 1st & 2nd & 3rd & avg. & & 1st & 2nd & 3rd & avg. & & \\
        \midrule

        \multicolumn{8}{l}{\textcolor{gray}{\textit{Commercial Video LLMs}}} \\
        Gemini-1.5-Pro~\citep{gemini} & - & 128 & 52.33 & 19.67 & 9.35 & 16.33  & 43.00  & 35.00 & 16.26 & 7.14 & 12.00  & 38.50  &   \xmark & \xmark \\
        GPT-4o~\citep{gpt4o} & -& 50 & 48.67 & 16.95 & 5.61 & 15.00  & 39.50  & 34.33 & 15.57 & 7.65 & 12.33  & 17.00  & \xmark  & \xmark\\ \midrule
        \multicolumn{8}{l}{\textit{\textcolor{gray}{Open-source Video LLMs}}} \\ 
        VideoChatGPT~\citep{videochatgpt} & LLaMA-7B& 100 & 35.33 & 4.7 & 1.87 &  3.33  & 33.50 & 18.00 & 3.11 & 0.51 &  3.00  & 3.50 & \xmark  & \xmark   \\
        VideoChat2~\citep{mvbench} & Vicuna-7B& 8 & 19.67 & 2.37 & 0.93 &  2.33  & 27.50 &  16.33 & 3.81 & 0.51 &  2.67 & 1.00 & \xmark  & \xmark   \\
        Video-LLaVA~\citep{videollava} & Vicuna-7B& 8 & 32.00 & 1.69 & 0.00 & 1.67  & 28.00 & 22.67 & 5.19 & 1.02 & 3.33  & 2.50 &  \xmark & \xmark  \\
        LLaMA-VID~\citep{llamavid} & Vicuna-7B& 128 &29.67 & 2.38 & 0.00 &  2.33 & 29.00 & 21.33 &3.80 &0.51 & 2.67 & 7.50 & \xmark  &  \xmark  \\
        MiniGPT4-Video~\citep{minigpt4video} & Mistral-7B& 45 & 25.00 & 4.75 & 1.87 &  4.00   & 23.00 & 12.67 & 2.08 & 0.51 &  1.67 & 3.00 & \xmark  & \xmark   \\ 
        PLLaVA~\citep{xu2024pllava} & Vicuna-7B& 16 & 37.33 & 3.73 & 0.93 &  3.33 & 30.00 & 21.00 & 3.46 & 0.00 & 1.33  & 3.00 & \xmark  &  \xmark  \\ 
        LLaVA-NeXT-Video~\citep{llavanextvideo} & Vicuna-7B& 32 & 30.33 & 2.37 & 0.93 & 3.00  & 30.50 & 17.00 & 2.08 & 0.51 & 2.00  & 1.50 & \xmark  & \xmark   \\
        ShareGPT4Video~\citep{sharegpt4video} & Llama3-8B& 16 &34.00 & 2.03 & 0.93 & 2.00  & 29.00 & 20.33 & 3.46 & 0.00 & 2.00  & 4.50  & \xmark  & \xmark   \\
        LLaMA-VID-13B~\citep{llamavid} & Vicuna-13B& 128 & 33.33 & 2.03 & 0.00 & 1.33  &  30.50 & 22.67 & 3.46 & 0.51 & 3.33  & 8.50  & \xmark  & \xmark   \\
        PLLaVA-13B~\citep{xu2024pllava} & Vicuna-13B& 16 & 41.33 & 3.39 & 0.00 &  2.67  & 25.00 & 25.67 & 5.54 & 2.04 & 4.33  & 6.50    & \xmark  & \xmark \\
        PLLaVA-34B~\citep{xu2024pllava} & Yi-34B& 16 & 29.00 & 4.07 & 0.00 &  3.67 & 28.50 & 18.67 & 4.50 &0.00  &  3.00 & 5.00  & \xmark  & \xmark   \\ 
        LLaVA-NeXT-Video-34B~\citep{llavanextvideo} & Yi-34B& 32 & 30.33 & 2.71 & 0.00 & 2.67  & 32.50 & 14.67 & 2.08 & 0.51 &  1.67 & 1.50  &   \xmark   & \xmark \\ \midrule
        LongVA~\citep{longva} & Qwen2-7B& 32 & 33.33 & 4.07 & 0.00 & 3.33  & 37.50 & 33.33 & 4.07 & 0.00 & 2.33  & 3.00  &  \xmark & \xmark  \\
        LongVILA~\citep{xue2024longvila} & Llama3-8B& 128 & 39.00  &  4.41  & 0.93 & 4.33  & 39.50 & 39.00  &  4.41  & 0.93 & 3.00  & 10.00  & \xmark  & \xmark  \\
        LongLLaVA~\citep{wang2024longllava} & Jamba-9B& 128 & 36.33 & 3.73 & 0.00 & 3.33  & 29.00  &36.33 & 3.73 & 0.00 & 3.67  & 10.00  &  \xmark & \xmark  \\
        VideoLLM-online~\citep{videollmonline} & Llama3-8B& 1 fps & 18.00 & 4.75 & 0.00 & 4.67  &  35.00 &  18.00 & 4.75 & 0.00 & 1.33  & 0.00  & 0.50  & \xmark \\
        VideoLLaMB~\citep{videollamb} & Vicuna-7B& 32 / 1 fps &32.67 & 2.71 & 0.00 & 2.33  & 29.50  & 32.67 & 2.71 & 0.00 & 3.00  & 3.00  & 0.00  & \xmark \\ 
        IXC2.5-OL~\citep{} & Qwen2-1.5B & 32 & 40.33  & 5.08  &  0.00  & 4.03 & 30.50  & 26.00 & 4.50 & 1.52 & 4.00  & 23.0 & \xmark & ? \\ \midrule
        
        \multicolumn{8}{l}{\textcolor{gray}{\textit{ OmniLLMs}}} \\
        
        VideoLLaMA2~\citep{videollama2} & Qwen2-7B & 8 & 41.00 & 12.88 & 0.00 & 10.33 &  35.00 & 23.33 & 4.15 & 0.51  &  3.00  & 5.00  & \xmark  & \xmark \\
        VITA~\citep{vita} & Mistrl-8$\times$7B & 16 & 8.67 & 0.00  & 0.00  & 0.00  & 39.00  & 11.33 & 3.11 & 1.52 & 2.00  & 1.50  & \xmark & 67.00 \\
        MiniOmini2~\citep{xie2024miniomni2} & Qwen2-0.5B & 1 & 17.00  & 5.08  &  0.93  & 4.67 & 14.00  & 6.00 & 1.00 & 0.00 & 1.00  & 1.00 & \xmark & \xmark \\

        \model(ours) & Qwen2-7B& 32 / 1 fps & 35.67 & 6.44 & 1.87 & 5.67   &  33.5 &  35.67 & 6.44 & 1.87 & 1.67 & 9.00  &  25.50 & 62.00  \\
        \model-a(ours) & Qwen2-7B& 32 / 1 fps & 28.33  & 2.37 & 0.00 & 2.00   & 13.00  & 19.33  & 3.11 & 0.51 & 3.00  &  7.50 &  1.50 &  68.5 \\
        
        \bottomrule
    \end{tabular}}
   
    \vspace{-.2in}
\end{table*}

\paragraph{Parallel Decoding}

We enhance inference speed to achieve \textit{real-time} interactions through parallel decoding \cite{lookahead, Zhao2024OuroborosGL, duplex, duo}. As illustrated in \cref{fig:parallel}, when the model is generating new tokens and a new input query arises, the model decodes the next token alongside the original token using a combination of causal masks, prefix masks, and block masks. Specifically, the causal mask is applied for the language model, the prefix mask pertains to the video context, and the block mask is designed to separate the decoding procedures of different queries in parallel. This method allows for the prediction of the next token while responding to a new input query in a single forward pass. To enhance the model's resilience to noise, once the probability of the next token for the new input query is obtained, we evaluate whether it constitutes noise. If it does, the token is removed from the KVCache, allowing the continuation of the previous generation process. If the query is deemed legitimate, we proceed to decode the new query with the video context, while masking the interrupted sequence being decoded. This approach enables the processing of new queries at any forward step, thereby maintaining low latency.

\paragraph{\data}

Building on the aforementioned framework, we crafted a small video-free synthetic instruction finetuning dataset, \textbf{\data},  with the assistance of GPT-4o. \data comprises four components: (i) the original instruction, which is a data replay from the instruction data of our base model, in our work, we use the LLaVA-NeXT~\cite{longva}; (ii) interleaved image-text instruction, which is created by reordering the question and image components of the original instruction; (iii) noise instruction, where GPT-4 is prompted to automatically generate statements that do not require a response; and (iv) stop instruction, where GPT-4 is prompted to generate stop phrases for the stop instruction.
Detailed descriptions regarding the instruction construction pipeline and prompts are provided in \ac{sm}. An overall cost of $\$4.91$ was incurred to construct the instruction dataset.

\section{Experiments}

\paragraph{Setup} We perform exhaustive evaluations of existing video-language models on \bench. We have meticulously selected three categories of baseline models for our analysis: commercial \acp{videollm}, open-source \acp{videollm}, and LAVLMs, which encompass a spectrum from visual to audio modalities. Within the open-source \acp{videollm}, we further explore models that vary in scalability, context length, and real-time design capabilities.

\subsection{Main Results}
\cref{tab:overall} presents the evaluation results. In summary, our analysis yields three key observations.
\begin{itemize}[leftmargin=*, noitemsep, topsep=0pt]
    \item \textbf{Challenges in Multi-Turn Tasks for Streaming Video.} In the context of multi-turn tasks such as State Grounding and the Multi-Turn Dependency task, models exhibit a notable decline in performance when required to handle more than a single state or reasoning step. When tasked with managing three states or reasoning steps, the majority of open-source models fail to accurately address all inquiries. These results underscore the limitations of current methodologies in managing dynamic environments and performing multi-turn reasoning, despite their demonstrated efficacy in static video scenarios.

    \item \textbf{Limitations in Audio-Visual Interaction.} Although our benchmark is explicitly designed to evaluate visual-audio interaction, current open-source models equipped with both visual and audio inputs do not outperform those with solely visual inputs. This discrepancy highlights a deficiency in the alignment of audio and visual features. Moreover, models with specialized speech training~\cite{vita,xie2024miniomni2} perform significantly worse than text input models, emphasizing the critical need for effective alignment and integration of multimodal inputs.

    \item \textbf{Model Size \vs Input Length.} Our experiments reveal that increasing model size does not necessarily enhance performance in streaming video tasks. Models with 7B parameters achieve performance levels comparable to those of larger models while maintaining greater efficiency. Conversely, models designed with long context capacities demonstrate improved performance in streaming tasks. Although we faced memory constraints with these models, we posit that balancing input length with memory efficiency is essential for effective understanding of streaming video content.
\end{itemize}

\subsection{Analysis}

\paragraph{Proactive Alerting}

We ablate \model with two backbones, Qwen2 and Llama3.1. Aside from accuracy, the precision and Intersection-over-Union (IoU) for all responses are computed throughout the entire video stream. We evaluate its \textbf{PA} ability under various settings, with results presented in \cref{tab:ground}. Further evaluation of the mixed model on the general video understanding task, VideoMME~\cite{videomme}, yields scores of $51.74$ for \model-Qwen2 and $43.52$ for \model-Llama3.1.

\begin{figure}[htbp!]
        \centering
   \includegraphics[width=\linewidth]{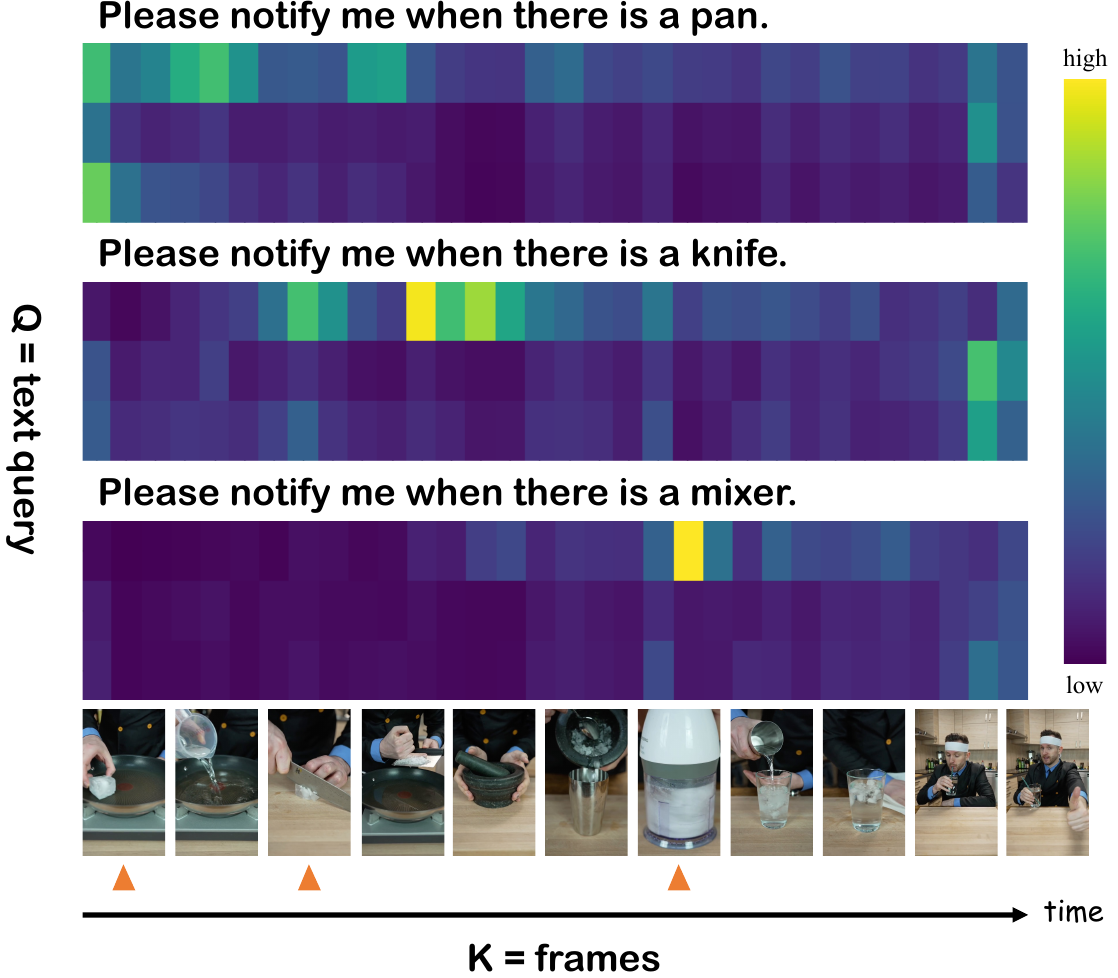}
       \vspace{-6mm}
   \caption{Attention feature map utilizes query as Q frames as K. The query consists of the last three tokens of the text query, while the key is represented by the mean-pooled frame.}
   \label{fig:grounding}
       \vspace{-.2in}
        \end{figure}

\noindent\textit{\underline{Findings}}\quad Our analysis reveals a significant performance gap between different LLMs.  These results are generally consistent with the performance in \cref{tab:overall}. We conclude that leveraging a model with strong general ability can enhance proactive capabilities without necessitating the construction of new data, which could potentially compromise the model's performance~\cite{videollmonline}. Moreover, interleaved data appear to improve the model's grounding ability, aligning with existing findings~\cite{mantis,llavainterleave}.  
Further investigation shows that the specific tokens in the query play a crucial role in achieving meaningful attention weights for grounding, where the tokens associated with the assistant role have demonstrated superior effectiveness.  
To effectively demonstrate the efficacy of our \model, we visualize the attention weights in \cref{fig:grounding}, in which a strong correlation is presented between the elements within the relevant frames and the input query, validating the effectiveness of our approach.

 \begin{table}[t!]
      \caption{Ablation Study Results for \model on the Proactive Alerting Task. ``interleave'': the tuning data comprising interleaved text and images. ``mix'': using \data. ``mean'': the attention weight is computed by averaging all tokens from the query.}
  \label{tab:ground}
  \vspace{-2mm}
  \resizebox{.95\linewidth}{!}{
  \begin{tabular}{lcccccc}
    \toprule
    \textbf{Method} & Precision & IoU & Accuracy \\ \midrule
    \model-Qwen & 31.60 & 13.90 & 25.00 \\
    \model-Qwen-interleave & 32.65 & 14.65  & 26.00 \\
    \model-Qwen-mix & 29.58 & 10.43  & 25.50 \\
    \model-Qwen-mix-mean & 5.50 & 0.00 &  6.50 \\ \midrule
    \model-Llama & 8.38 & 2.47 & 10.50 \\
    \model-Llama-interleave & 9.17  & 1.05 & 10.00 \\
    \model-Llama-mix & 10.63 & 5.26 & 11.50  \\
    \model-Llama-mix-mean & 0.50 & 0.00 & 0.50 \\
    
    \bottomrule
  \end{tabular}
  }
  \vspace{-.13in}

    \end{table}

    \begin{table}[t!]
  \centering
    \caption{Performance on general video understanding task from VideoMME~\cite{videomme}.}
  \label{tab:general}
  \vspace{-2mm}

  \resizebox{.95\linewidth}{!}{
  \begin{tabular}[\linewidth]{@{}lcccc@{}}
    \toprule
    Model & Short & Medium & Long & General  \\\midrule
    LongVA & 61.1 & 48.3 & 45.4 & 52.4  \\
    LongVA (DataReplay) & 60.9 & 50.7 & 45.0 & 52.2  \\
    \model (Interleave) & 60.3 & 50.6 & 43.9 & 51.6  \\
    \model (Noise) & 60.3 & 51.4 & 45.7 &  52.5  \\
    \model (Stop) & 60.3 & 60.8  & 44.0 & 51.7  \\
    \model & 60.6 & 50.8 & 43.9 & 51.7 \\
    
    \bottomrule
  \end{tabular}
  }
  \vspace{-.16in}
\end{table}

    \begin{figure}[htbp!]
    \centering
    \includegraphics[width=\linewidth]{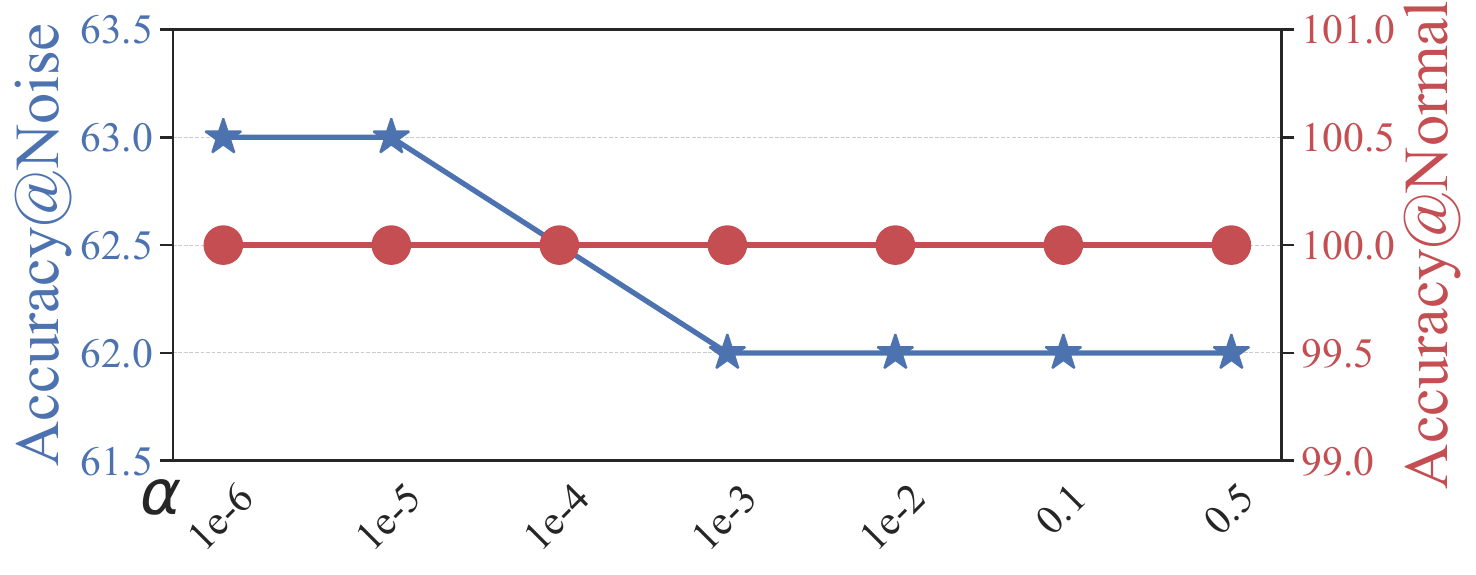}
    \vspace{-8mm}
    \caption{Performance on the Proactive Turn-taking task for noise and normal query over different scaling factor.}
    \label{fig:proactive-turn}
        \vspace{-.1in}
    \end{figure}

\paragraph{Proactive Turn-taking}


To evaluate the efficacy of the instruction data, we applied the proactive turn-taking task to the \model, which was finetuned on \data. 

\noindent\textit{\underline{Findings}}\quad The results in \cref{fig:proactive-turn} indicate that, after tuning with \data, our method successfully handled all legitimate queries. When subjected to noise input queries, the model demonstrated resilience across a broad range of the hyperparameter $\alpha$. This suggests that our proposed instructional data effectively facilitates learning the format without compromising the model's performance on standard queries.

\paragraph{Influence on General Task}

To evaluate the effectiveness of our proposed \model on general tasks, we further examine the model's capability using the VideoMME benchmark for general video understanding. The results are listed in \cref{tab:general}, where each configuration was trained on an identical dataset to ensure a fair comparison.

\noindent\textit{\underline{Findings}}\quad Our findings suggest that the introduction of an interleaved image-text instruction format and stop instructions has the most significant impact on the results, primarily due to the heterogeneous nature of the data format. When all these instructional data types are combined, there is a compromise in performance; however, it still surpasses the outcomes of the previously mentioned individual instruction types. This mixture achieves an effective balance, transitioning from general MLLM to interactive \acp{videollm} without training on any video.

\section{Conclusion}

In this work, we introduce \bench, which evaluates the interactive capabilities of systems processing streaming video in open-world contexts. \bench addresses challenges like streaming temporal state awareness and proactive reasoning with turn-taking. To advance real-time interactive reasoning, we propose a novel framework, \model, which enhances proactive turn-taking and efficient streaming capabilities. Our evaluations of previous MLLMs reveal significant limitations in handling multi-turn tasks and modality alignment. As such, we call for future research on efficient designs for open-world interactive \acp{omnillm}. \\
\textbf{Acknowledgments}\quad The authors thank the reviewers for their insightful suggestions on improving the manuscript. This work presented herein is supported by the National Natural Science Foundation of China (62376031). 


\clearpage

{
    \small
    \bibliographystyle{ieeenat_fullname}
    \bibliography{main}
}

\clearpage
\setcounter{page}{1}
\maketitlesupplementary
\appendix
\section{Audio Adaption Analysis}
\label{supp:audio}

\begin{table*}[t!]
        
    \caption{\textbf{Performance comparison of existing OmniLLM on \bench}.  The 1st, 2nd, 3rd of \textbf{SG} and \textbf{MD} tasks represent the cumulative accuracy up to and including these stages. The ``avg.'' indicates average accuracy across all data points. } 
    \label{tab:audio}
    \centering
    \resizebox{\linewidth}{!}{
    \begin{tabular}{l|c|c|ccccccccc|ccc}
        \toprule
        \multirow{2}{*}{\textbf{Models}} & \multirow{2}{*}{\textbf{LLM}} & \multirow{2}{*}{\makecell{\textbf{Num} \\ \textbf{Frames}}} & \multicolumn{4}{c}{\textbf{SG}} & \multirow{2}{*}{\textbf{AP}} & \multicolumn{4}{c|}{\textbf{MD}} & \multirow{2}{*}{\textbf{SI}} & \multirow{2}{*}{\textbf{PA}} & \multirow{2}{*}{\textbf{PT}}  \\ 
        \cmidrule(lr){4-7} \cmidrule(lr){9-12}
        & & & 1st & 2nd & 3rd & avg. & & 1st & 2nd & 3rd & avg. & & \\
        \midrule

        \multicolumn{8}{l}{\textcolor{gray}{\textit{Commercial Video LLMs}}} \\
        Gemini-1.5-Pro~\citep{gemini} & - & 128 & 52.33 & 19.67 & 9.35 & 16.33  & 43.00  & 35.00 & 16.26 & 7.14 & 12.00  & 38.50  &   \xmark & \xmark \\
        GPT-4o~\citep{gpt4o} & -& 50 & 48.67 & 16.95 & 5.61 & 15.00  & 39.50  & 34.33 & 15.57 & 7.65 & 12.33  & 17.00  & \xmark  & \xmark\\ \midrule
        {\textcolor{gray}{\textit{ OmniLLMs}}} \\
        
        VideoLLaMA2~\citep{videollama2} & Qwen2-7B & 8 & 41.00 & 12.88 & 0.00 & 10.33 &  35.00 & 23.33 & 4.15 & 0.51  &  3.00  & 5.00  & \xmark  & \xmark \\
        VITA~\citep{vita} & Mistrl-8$\times$7B & 16 & 8.67 & 0.00  & 0.00  & 0.00  & 39.00  & 11.33 & 3.11 & 1.52 & 2.00  & 1.50  & \xmark & 67.00 \\
        MiniOmini2~\citep{xie2024miniomni2} & Qwen2-0.5B & 1 & 17.00  & 5.08  &  0.93  & 4.67 & 14.00  & 6.00 & 1.00 & 0.00 & 1.00  & 1.00 & \xmark & \xmark \\ \midrule
        
        \model(ours) & Qwen2-7B& 32 / 1 fps & 35.67 & 6.44 & 1.87 & 5.67   &  33.5 &  35.67 & 6.44 & 1.87 & 1.67 & 9.00  &  25.50 & 62.00  \\
        \model-a(ours) & Qwen2-7B& 32 / 1 fps & 28.33  & 2.37 & 0.00 & 2.00   & 13.00  & 19.33  & 3.11 & 0.51 & 3.00  &  7.50 &  1.50 &  68.5 \\
        
        \bottomrule
    \end{tabular}}
   
    \vspace{-.2in}
\end{table*}

In this section, we further explore the adaptation of our methods to audio speech input. To adapt \model\ to receive audio queries, we fine-tuned it on a randomly selected subset of the VoiceAssistant dataset \cite{xie2024miniomni}, which comprises 30,000 audio instructions. To ensure a fair comparison, we maintained the same hyperparameters and other settings as those used in the tuning of \model. The results are presented in Table~\ref{tab:audio}. Our findings indicate that tuning the model on purely audio instruction data, without incorporating visual data, does not enhance its proactive turn-taking ability. Consequently, we converted the queries in \data\ to speech using CosyVoice and mixed them with the VoiceAssistant subset used during the tuning of \model-a. After integrating this audio data, we achieved a score of $68.5$ on the \textbf{PT} task. Overall, the introduction of audio instruction data still limits the performance of tasks requiring both visual and audio inputs. We believe this limitation arises from the lack of mixed visual and audio data during the training phase. In future work, we aim to enhance the model's audio understanding capabilities by incorporating more high-quality multimodal data.

\section{Highlight Spot Algorithm}
\label{supp:hightlight}

\begin{algorithm}[ht]
\caption{Highlight Spot}
\label{algo:hightlight}
\begin{algorithmic}[1]
    \Require Video stream $V_{\infty}$, query $q$, threshold $\gamma$, Gaussian factor $\alpha$
    \State \texttt{highlight\_spot.init()} 
    \ForAll{frame $v$ in $V_\infty$}
        \State \texttt{KVCache}.update($W_Kv$, $W_Vv$)
        \State $\texttt{attn} \gets \texttt{SelfAttn}(v \oplus q, \texttt{KVCache})$
        \State $(\mu, \sigma) \gets \texttt{std\_mean}(\texttt{attn})$
        \State $\delta \gets \mu + \alpha \times \sigma$
        \State $\texttt{cands} \gets \{t ; \texttt{attn}[t] > \delta\}$
        \ForAll{$c_t$ in \texttt{cands}}
            \State $c_t \gets \texttt{highlight\_spot.get}(t) + 1$
            \State $\texttt{highlight\_spot.update}(i, c_i)$
        \EndFor
        \If{\texttt{highlight\_spot.heap} \textbf{is not empty}}
            \State $(i, c) \gets \texttt{highlight\_spot.peek()}$
            \If{$freq > \gamma$}
                \State \texttt{send}($i$)
            \EndIf
        \EndIf
    \EndFor
\end{algorithmic}
\end{algorithm}

In this section, we present the pseudo-code of our proposed training-free highlight spot algorithm, as illustrated in Algorithm~\ref{algo:hightlight}. For any transformer-based model, incoming streaming video frames are stored in the KVCache to avoid redundant computations. Subsequently, we compute the attention weights from the model's final layer using the text query as the key and the video as the value. We then identify and save the frame indices whose attention weights exceed a threshold, determined by the mean and variance of the previous attention weights. These indices are labeled as consistently salient frames, signifying the frames that need to be highlighted. The consistency threshold is a hyperparameter, which is set to $4$ in our experiments. Furthermore, we introduce an initial latency step to mitigate the challenges associated with calculating the mean and variance; in practice, this latency step is set to $2$.

\section{Single Question Analysis of Multi-turn Dependency Reasoning}
\label{supp:split-dependency}

\begin{table}[h]
  \centering
  \caption{Multi-turn Dependency Reasoning}
  \resizebox{\linewidth}{!}{
  \begin{tabular}{@{}lcccc@{}}
    \toprule
    Models & Step=1 & Step=2 & Step=3 & Overall\\
    \midrule
    \multicolumn{5}{l}{\textcolor{gray}{\textit{ Commercial Video LLMs}}} \\
    Gemini-1.5-Pro & 52.33 & 34.24 & 36.45 & 16.33  \\
    GPT-4o & 48.67 & 31.53 & 20.56 &  15.00 \\\midrule
    \multicolumn{5}{l}{\textcolor{gray}{\textit{ Open-source Video LLMs}}} \\
    VideoChatGPT & 18.00 & 13.49 & 11.22  & 3.00 \\
    VideoChat2 & 16.33 & 13.15 & 12.24 & 2.67  \\
    Video-LLaVA & 22.67 & 13.49 & 16.33 & 3.33  \\
    LLaMA-VID &21.33 &15.22 &13.78 &2.67   \\
    MiniGPT4-Video & 12.67 & 6.57 & 8.67 & 1.67   \\
    PLLaVA & 21.00 & 13.49 & 17.35 & 1.33  \\
    LLaVA-NeXT-Video & 17.00 & 10.03 & 10.71 & 2.00   \\
    ShareGPT4Video & 20.33 & 15.57 & 14.80 & 2.00  \\
    LLaMA-VID-13B & 22.67 & 14.88 & 14.29 & 3.33   \\
    PLLaVA-13B & 25.67 & 17.80 & 16.84 & 4.33  \\
    PLLaVA-34B & 18.67 & 17.30 & 10.20 &3.00   \\
    LLaVA-NeXT-Video-DPO-34B & 14.67 & 14.53 & 12.24 & 1.67   \\\midrule
    LongVA & 20.67 & 16.27 & 13.78 & 2.33   \\
    LongVILA & 22.33 & 14.19 & 14.29 & 3.00   \\
    LongLLaVA & 26.33 & 18.69 & 20.41 & 3.67   \\
    VideoLLM-online& 11.67 & 7.27 & 10.71 & 1.33   \\
    VideoLLaMB & 18.67 & 13.15 & 17.86 &  3.00  \\\midrule
    
    \multicolumn{5}{l}{\textcolor{gray}{\textit{ OmniLLMs}}} \\
    VideoLLaMA2 & 23.33 & 15.92 & 18.78 & 5.00  \\
    VITA & 11.33 & 12.80 & 8.63 & 2.00   \\
    MiniOmini2 & 6.00 & 3.11 & 2.03 & 1.00   \\\midrule
    \model & 19.33 & 10.73 & 12.18 & 1.67   \\
    \bottomrule
  \end{tabular}}
  
  \label{tab:split-dependency}
\end{table}

In this section, we detail the accuracy of each step in the multi-turn dependency reasoning task. The results are presented in Table~\ref{tab:split-dependency}. Unlike the results presented in Table~\ref{tab:overall}, this experiment focuses solely on the accuracy of individual reasoning steps. Generally, we observe a decline in accuracy as the number of steps increases. However, in certain instances, accuracy at a later step exceeds that of a previous one. We attribute this anomaly to potential hallucinations generated by the language models. Overall, there is a significant drop in accuracy across successive steps, underscoring the importance of multi-step reasoning in evaluation. This approach helps to mitigate errors introduced by language models, demonstrating the necessity of a step-by-step evaluation process.

\section{Single Question Analysis of Dynamic State Grounding}
\label{supp:split-transition}

\begin{table}[h]
  \centering
  \caption{Dynamic State Grounding}
  \resizebox{\linewidth}{!}{
  \begin{tabular}{@{}lcccc@{}}
    \toprule
    Models & State=1 & State=2 & State=3 & Overall\\
    \midrule
    \multicolumn{5}{l}{\textcolor{gray}{\textit{ Commercial Video LLMs}}} \\
    Gemini-1.5-Pro & 35.00 & 37.02 & 38.78 & 12.00  \\
    GPT-4o & 34.33 & 33.56 & 37.24   & 12.33 \\\midrule
    \multicolumn{5}{l}{\textcolor{gray}{\textit{ Open-source Video LLMs}}} \\
    VideoChatGPT & 35.33 & 17.97 & 10.28 & 3.33 \\
    VideoChat2 & 19.67 & 14.23 & 6.54 & 2.33  \\
    Video-LLaVA & 32.00 & 16.27 & 11.21 & 1.67   \\
    LLaMA-VID & 29.67 & 13.56 & 7.48 & 2.33   \\
    MiniGPT4-Video & 25.00 & 15.25 & 14.02 & 4.00  \\
    PLLaVA & 37.33 & 13.56 & 10.29 & 3.33  \\
    LLaVA-NeXT-Video & 30.33 & 12.20 & 6.54 & 3.00   \\
    ShareGPT4Video & 34.00 & 13.22 & 10.28 & 2.00  \\
    LLaMA-VID-13B & 33.33 & 14.24 & 6.54 & 1.33   \\
    PLLaVA-13B & 41.33 & 13.90 & 12.15 & 2.67   \\
    PLLaVA-34B & 29.00 & 14.24 & 10.28 & 3.67  \\
    LLaVA-NeXT-Video-DPO-34B & 30.33 & 11.19 & 5.61 &  2.67 \\\midrule
    LongVA & 33.33 & 15.59 & 8.41 & 3.33  \\
    LongVILA & 39.00  &  16.95 & 14.02  & 4.33  \\
    LongLLaVA & 36.33 & 11.53 & 7.48 & 3.33   \\
    VideoLLM-online & 18.00 & 13.56 & 5.61 & 4.67   \\
    VideoLLaMB & 32.67 & 14.58 & 10.28 & 2.33  \\\midrule
    \multicolumn{5}{l}{\textcolor{gray}{\textit{ Open-source Video LLMs}}} \\
    VideoLLaMA2 & 41.00 & 26.78 & 10.28 &  10.33  \\
    VITA & 8.67 & 8.14 & 2.80 & 0.00  \\
    MiniOmini2 & 17.00 & 14.92 & 10.28 & 4.67   \\\midrule
    
    \model &  35.67& 13.22 & 6.54 & 5.67   \\
    \bottomrule
  \end{tabular}}
  
  \label{tab:split-transition}
\end{table}

In this section, we extend our analysis of the Dynamic State Grounding task by examining the performance on each individual question. The results, as detailed in Table~\ref{tab:split-transition}, indicate a notable decline in performance as the number of states increases. This decline can be attributed to the increased length of the video context and dialogue history, which complicates the process of dynamically grounding the current state to derive the correct answer. Furthermore, our analysis did not reveal significant performance differences across different models at the initial state. However, the performance gap widens as the number of states increases, underscoring the importance of a model's ability to handle longer contexts while maintaining effective grounding capabilities.

\section{Annotation Details}


\subsection{Raw Video Data Collection}
To enhance our dataset, we specifically collect data from YouTube, concentrating primarily on videos that are particularly commonly useful in our real-life. We also focus on the videos which are in content involving personal introductions and interpersonal interactions. 

\subsection{Annotation Tool}

\begin{figure}[h]
    \centering
    \includegraphics[width=\linewidth]{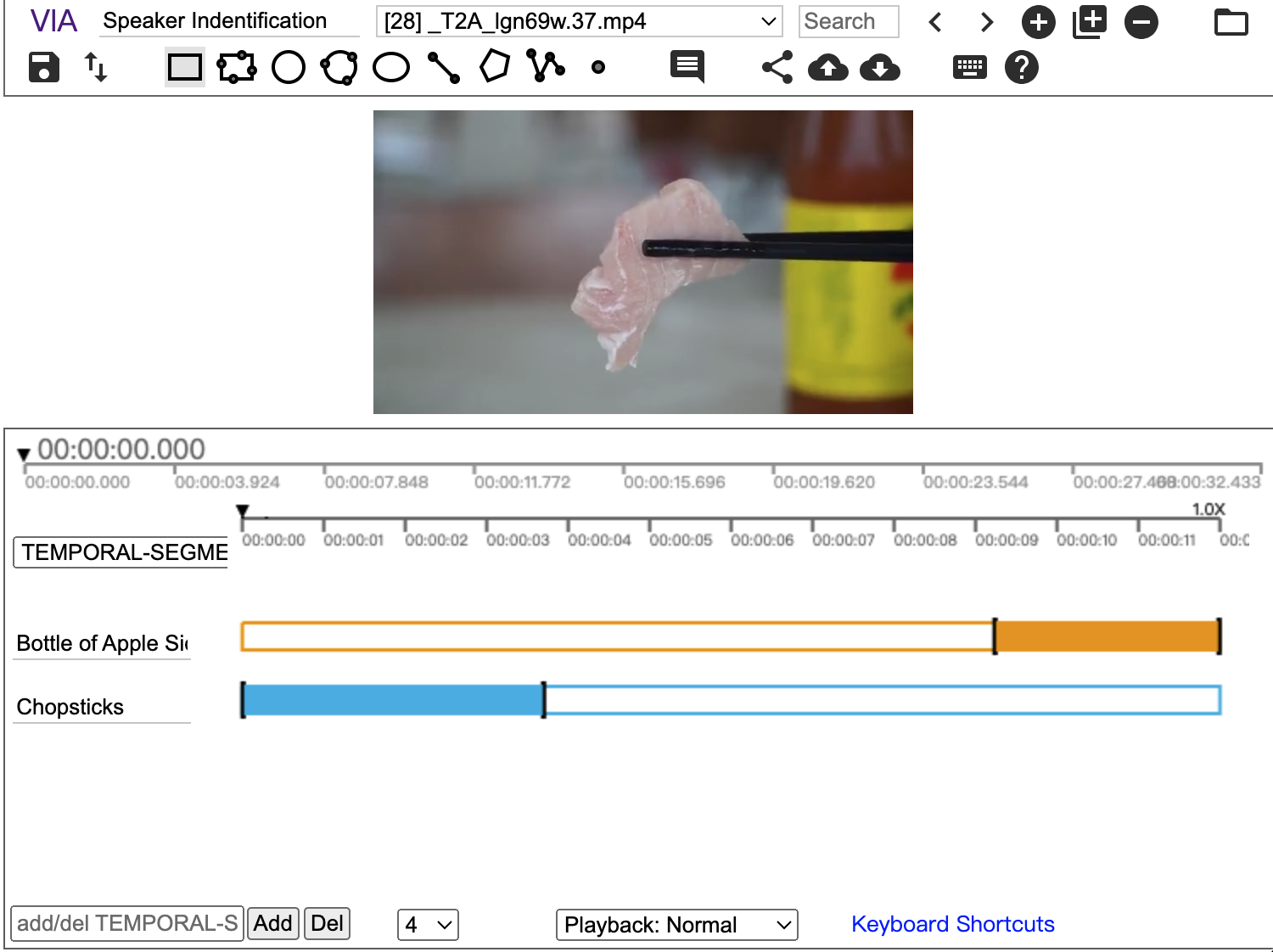}
    \caption{The Front-End Interface for Human Annotation}
    \label{fig:labelfont}
\end{figure}

The front-end interface for human annotation is depicted  in Figure~\ref{fig:labelfont}. In this interface, each question or statement is associated with the most relevant time span, which serves either as part of the label or as an aid for subsequent annotation tasks.

\subsection{Annotation Guidelines}

\begin{table}[h]
    \centering
    \resizebox{\linewidth}{!}{
    \begin{tabular}{cc}
    \toprule
         \textbf{Category} & \textbf{Question} \\
            \hline
            Object State & How many objects are in the scene? \\
            & How many people are in the room? \\
            & What is the color of the car? \\
            & Is the door open or closed? \\
            \hline
            Spatial Relations & Where is the cat relative to the chair? \\
            \hline
            Dynamic Spatial Relations & Is the person walking towards or away from the camera? \\
            & Where is the ball relative to the player? \\
            \hline
            Action State & What is the person doing? \\
            & What activity is happening in the scene? \\
            \hline
            Scene State & Is the room well-lit or dim? \\
            & What is the weather like? \\
            & Is the street busy or quiet? \\
            & What is the context of the scene? \\
            \hline
            Human Object Interaction & Is the person holding the book? \\
            \hline
            Human Human Interaction & Are the two people shaking hands? \\
            & What is the interaction between the two characters? \\
            \hline
            Group Dynamics & How are the group members interacting? \\
            \hline
            Emotional State & What is the person's emotional state? \\
            \hline
            Audio/Speech State & What does the speaker mentioned? \\\bottomrule
    \end{tabular}}
    \caption{Annotation hints for annotators including category and example question.}
    \label{tab:labelguide}
\end{table}

To ensure that annotators produce high-quality annotations that align with our specified standards, we provide detailed guidelines, including examples of various question types. The list of hints is demonstrated in Table~\ref{tab:labelguide}.

\section{\data Construction Details}

\subsection{Noise data prompt}

We employ GPT-4o to autonomously generate noise data for the purpose of instruction tuning. The prompt utilized for the generation of noise data is detailed below.
\begin{tcolorbox}[colback=gray!10, colframe=gray!50, width=\linewidth, sharp corners]
\textit{\color{black}
You are a sophisticated AI designed to simulate human-like conversation by generating 'noise.' This noise consists of naturally flowing statements that mimic the user's perspective. ------
Review the user's questions and the assistant's responses carefully. Using this information, create coherent declarative statements that reflect the user's voice. These should resemble everyday human dialogue and do not require a response from the assistant.
Ensure your output is in the form of declarative sentences and avoid questions.
Keep the noise brief and in casual, conversational English. But do NOT need response}
\end{tcolorbox}

\subsection{Stop Words}

We compile a set of frequently used stop words to incorporate into our instructional data, thereby serving as the designated stop words: 
``That's a good point, and'', ``Let me stop you there'', ``Just a second'', ``I don't mean to be rude, but'', ``If I could interject'', ``Pardon me, but'', ``Sorry to interrupt'', ``Before you continue'', ``Can we pause for a moment?'', ``May I add something here?'', ``I apologize for cutting in'', ``Could I stop you for a second?'', ``I’d like to add'', ``Could I clarify something?'', ``I have a quick question'', ``This reminds me of'', ``Let me add to that'', ``Can I share my thoughts?'', ``Hold on a moment'', ``One moment, please'', ``Allow me to explain'', ``Excuse me'', ``Can I jump in for a moment?'', ``I see what you mean, but'', ``I think it’s important to mention''

\section{\model Implementation Details}

\begin{table}[h]
\begin{center}
\resizebox{0.6\linewidth}{!}{
\begin{tabular}{lccc}
\toprule
\multicolumn{1}{l}{\bf Hyperparam} &\multicolumn{1}{c}{\bf \model} \\ 
\midrule 
$\alpha$ & 2 \\
$\beta$ & 0.2 \\
$\gamma$ & 4 \\
Model Max Length & 32000 \\
Learning Rate & 1e-5 \\
Warmup Ratio & 0.03 \\
Per Device Batch Size & 1  \\
Gradient Accumulation Steps & 4 \\
Epoch & 1 \\
\bottomrule
\end{tabular}}
\caption{Hyperparameters for \model.}
\label{tab:hyper}
\end{center}
\end{table}

In practice, we conduct the training process using four Nvidia A800 GPUs, which requires approximately one hour to fine-tune the model. Table~\ref{tab:hyper} presents a detailed account of the hyperparameters employed during both the training and inference procedures.

\end{document}